%% file: gcl_adv.tex
\documentclass{article}

\usepackage[final]{nips_2016}

\usepackage[utf8]{inputenc} 
\usepackage[T1]{fontenc}    
\usepackage{hyperref}       
\usepackage{url}            
\usepackage{booktabs}       
\usepackage{amsfonts}       
\usepackage{nicefrac}       
\usepackage{microtype}      

\usepackage{graphics} 
\usepackage{epsfig} 
\usepackage{mathptmx} 
\usepackage{times} 
\usepackage{amsmath} 
\usepackage{amssymb}  
\usepackage{verbatim}
\usepackage{algorithm}
\usepackage{algorithmic}
\usepackage{color}
\usepackage{wrapfig}
\usepackage{multirow}

\usepackage{mathtools}
\usepackage{stmaryrd}
\DeclareMathAlphabet{\mathcal}{OMS}{cmsy}{m}{n}

\include{defs}

\DeclarePairedDelimiter{\bracketdelims}{[}{]}
\newcommand{\brackets}[1]{\bracketdelims*{#1}}
\newcommand{\E}[2]{\mathbb{E}_{#1}\brackets{#2}}

\DeclarePairedDelimiter{\pdelims}{(}{)}
\newcommand{\of}[1]{\pdelims*{#1}}
\newcommand{\logo}[1]{\log\of{#1}}
\newcommand{\expo}[1]{\exp\of{#1}}
\newcommand{\mix}{\mu}
\renewcommand{\c}{\cost_\params}
\newcommand{\mixeddensity}[1]{\frac 1Z \exp\of{-\c\of{#1}} + q\of{#1}}
\newcommand{\expc}{\expo{-\c\of{\traj}}}
\newcommand{\mut}{\widetilde{\mu}\of{\traj}}
\newcommand{\loss}[2]{\obj_{\text{#1}}\of{#2}}
\newcommand{\qt}{q\of{\traj}}
\newcommand{\pt}{p_{\theta}\of{\traj}}
\newcommand{\pebm}{p_{\theta}\of{\gansamp}}

\title{A Connection Between Generative Adversarial Networks, Inverse Reinforcement Learning, and Energy-Based Models
}

\author{Chelsea Finn$^*$, Paul Christiano\thanks{~Indicates equal contribution.}~, Pieter Abbeel, Sergey Levine\\
University of California, Berkeley\\
\texttt{\{cbfinn,paulfchristiano,pabbeel,svlevine\}@eecs.berkeley.edu}\\
}

\begin{document}

\maketitle

\begin{abstract}
Generative adversarial networks (GANs) are a recently proposed class of generative models in which a generator is trained to optimize a cost function that is being simultaneously learned by a discriminator. While the idea of learning cost functions is relatively new to the field of generative modeling, learning costs has long been studied in control and reinforcement learning (RL) domains, typically for imitation learning from demonstrations. In these fields, learning the cost function underlying observed behavior is known as inverse reinforcement learning (IRL) or inverse optimal control.
While at first the connection between cost learning in RL and cost learning in generative modeling may appear to be a superficial one,
we show in this paper that certain IRL methods are in fact mathematically equivalent to GANs. In particular, we demonstrate an equivalence between a sample-based algorithm for maximum entropy IRL and a GAN in which the generator's density can be evaluated and is provided as an additional input to the discriminator. Interestingly, maximum entropy IRL is a special case of an energy-based model. We discuss the interpretation of GANs as an algorithm for training energy-based models, and relate this interpretation to other recent work that seeks to connect GANs and EBMs. By formally highlighting the connection between GANs, IRL, and EBMs, we hope that researchers in all three communities can better identify and apply transferable ideas from one domain to another, particularly for developing more stable and scalable algorithms: a major challenge in all three domains.
\end{abstract}

\section{Introduction}
\label{introduction}

Generative adversarial networks (GANs) are a recently proposed class of generative models in which a generator is trained to optimize a cost function that is being simultaneously learned by a discriminator~\cite{gpmxw-gan-14}. While the idea of learning objectives is relatively new to the field of generative modeling, learning cost or reward functions has long been studied in control~\cite{bv-co-04} and was popularized in 2000 for reinforcement learning problems~\cite{nr-airl-00}. In these fields, learning the cost function underlying demonstrated behavior is referred to as inverse reinforcement learning (IRL) or inverse optimal control (IOC). At first glance, the connection between cost learning in RL and cost learning for generative models may appear to be superficial; however, if we apply GANs to a setting where the generator density can be efficiently evaluated, the result is exactly equivalent to a sample-based algorithm for maximum entropy (MaxEnt) IRL.
Interestingly, as MaxEnt IRL is an energy-based model, this connection suggests a method for using GANs to train a broader class of energy-based models.

MaxEnt IRL is a widely-used objective for IRL, proposed by Ziebart et al.~\cite{zmbd-meirl-08}. Sample-based algorithms for performing maximum entropy (MaxEnt) IRL have scaled cost learning to scenarios with unknown dynamics, using nonlinear function classes, such as neural networks~\cite{bkp-reirl-11,kprs-lofm-13,fla-gcl-16}. We show that the gradient updates for the cost and the policy in these methods can be viewed as the updates for the discriminator and generator in GANs, under a specific form of the discriminator. The key difference to a generic discriminator is that we need to be able evaluate the density of the generator, which we integrate into the discriminator in a natural way.

Traditionally, GANs are used to train generative models for which it is not possible to evaluate the density. When it is possible to evaluate the density, for example in an autoregressive model, it is typical to maximize the likelihood of the data directly. By considering the connection to IRL, we find that GAN training may be appropriate even when density values are available. For example, suppose we are interested in modeling a complex multimodal distribution, but our model does not have enough capacity to represent the distribution. Then maximizing likelihood will lead to a distribution which ``covers'' all of the modes, but puts most of its mass in parts of the space that have negligible density under the data distribution. These might be images that look extremely unrealistic, nonsensical sentences, or suboptimal robot behavior. A generator trained adversarially will instead try to ``fill in'' as many of modes as it can, without putting much mass in the space between modes. This results in lower diversity, but ensures that samples ``look like'' they could have been from the original data.

By drawing an exact correspondence between adaptive, sample-based algorithms for MaxEnt IRL and GAN training, we show that this phenomenon occurs and is practically important: GAN training can significantly improve the quality of samples even when the generator density can be exactly evaluated. This is precisely analogous to the observed ability of inverse reinforcement learning to imitate behaviors that cannot be successfully learned through behavioral cloning~\cite{rgb-rilsp-11}, direct maximum likelihood regression to the demonstrated behavior.

Interestingly, the maximum entropy formulation of IRL is a special case of an energy-based model (EBM)~\cite{z-mpabp-10}. The learned cost in MaxEnt IRL corresponds to the energy function, and is trained via maximum likelihood. Hence, we can also show how a particular form of GANs can be used to train EBMs.
Recent works have recognized a connection between EBMs and GANs~\cite{kb-ddgm-16,zml-ebgan-16}. In this work, we particularly focus on EBMs trained with maximum likelihood, and expand upon the connection recognized by Kim \& Bengio~\cite{kb-ddgm-16} for the case where
the generator's density can be computed.
By formally highlighting the connection between GANs, IRL, and EBMs, we hope that researchers in all three areas can better identify and apply transferable ideas from one domain to another.

\vspace{-0.2cm}
\section{Background}
\label{sec:background}
\vspace{-0.2cm}

In this section, we formally define generative adversarial networks (GANs), energy-based models (EBMs), and
inverse reinforcement learning (IRL), and introduce notation.

\subsection{Generative Adversarial Networks}\label{gans}

Generative adversarial networks are an approach to generative modeling where two models are trained simultaneously: a generator $G$ and a discriminator $D$.
The discriminator is tasked with classifying its inputs as either the output of the generator, or actual samples from the underlying data distribution $\datadistr$.
The goal of the generator is to produce outputs that are classified by the discriminator as coming from the underlying data distribution~\cite{gpmxw-gan-14}.

Formally,
the generator takes noise as input and outputs a sample $\gansamp \sim G$, while
the discriminator takes as input a sample $\gansamp$ and outputs the probability $D(\gansamp)$ that the sample was from the data distribution.
The discriminator's loss is the average log probability it assigns to the correct classification, evaluated on an equal mixture of real samples and outputs from the generator:
\[\loss{discriminator}{D} = \E{\gansamp \sim p}{- \log{D(\gansamp)}} + \E{\gansamp \sim G}{- \logo{1-D(\gansamp)}}.\]
The generator's loss can be defined one of several similar ways.
The simplest definition, originally proposed in \cite{gpmxw-gan-14}, is simply the opposite of the discriminator's loss.
However, this provides very little training signal if the generator's output can be easily distinguished from the real samples.
It is common to instead use the log of the discriminator's confusion~\cite{gpmxw-gan-14}.
We will define the generator's loss as the sum of these two variants:
\[\loss{generator}{G} = \E{\gansamp \sim G}{- \log{D(\gansamp)}} + \E{\gansamp \sim G}{ \logo{1-D(\gansamp)}}.\]

\subsection{Energy-Based Models}
\label{sec:ebms}

Energy-based models~\cite{lchr-ebl-06} associate an energy value $E_\params(\gansamp)$ with a sample $\gansamp$, modeling the data as a Boltzmann distribution:
\vspace{-0.1cm}
\begin{equation}
\pebm = \frac 1Z \expo{-E_\params(\gansamp)} 
\end{equation}
The energy function parameters $\params$ are often chosen to maximize the likelihood of the data; the main challenge in this optimization is evaluating the partition function $Z$, which is an intractable sum or integral for most high-dimensional problems. A common approach to estimating $Z$ requires sampling from the Boltzmann distribution $\pebm$ within the inner loop of learning. 

Sampling from $\pebm$ can be approximated by using Markov chain Monte Carlo (MCMC) methods; however, these methods face issues when there are several distinct modes of the distribution and, as a result, can take arbitrarily large amounts of time to produce a diverse set of samples.
Approximate inference methods can also be used during training, though the energy function may incorrectly assign low energy to some modes if the approximate inference method cannot find them~\cite{lchr-ebl-06}.

\subsection{Inverse Reinforcement Learning}

The goal of inverse reinforcement learning is to infer the cost function underlying demonstrated behavior~\cite{nr-airl-00}. It is typically assumed
that the demonstrations come from an expert who is behaving near-optimally under some unknown
cost. In this section, we discuss MaxEnt IRL and guided cost learning, an algorithm for MaxEnt IRL.

\subsubsection{Maximum entropy inverse reinforcement learning}

Maximum entropy inverse reinforcement learning models the demonstrations using a Boltzmann distribution,
where the energy is given by the cost function $\cost_\params$:
$$
\pt = \frac{1}{Z}\exp(-\cost_\params(\traj)),\label{eq:ioc_obj}
$$
Here, $\traj = \{\state_1,\action_1,\dots,\state_T,\action_T\}$ is a trajectory; $\cost_\params(\traj) = \sum_t \cost_\params(\state_t,\action_t)$ is a learned cost
function parametrized by $\params$; $\state_t$ and $\action_t$ are the state and action at time step $t$; and the partition function $Z$ is the integral of $\expc$ over all trajectories that are consistent with the environment dynamics.\footnote{This formula assumes that $\state_{t+1}$ is a deterministic function of the previous history. A more general form of this equation can be derived for stochastic dynamics~\cite{z-mpabp-10}. However, the analysis largely remains the same: the probability of a trajectory can be written as the product of conditional probabilities, but the conditional probabilities of the states $\state_t$ are not affected by $\theta$ and so factor out of all likelihood ratios.}

Under this model, the optimal trajectories have the highest likelihood, and the expert can generate suboptimal trajectories with a probability that decreases exponentially as the trajectories become more costly.
As in other energy-based models, the parameters $\theta$
are optimized to maximize the likelihood of the demonstrations.
Estimating the partition function $Z$ is difficult for large or continuous domains, and presents the main computational challenge.
The first applications of this model computed $Z$ exactly with dynamic programming \cite{zmbd-meirl-08}.
However, this is only practical in small, discrete domains, and is impossible in domains where the system dynamics $p(\state_{t+1} | \state_t, \action_t)$ are unknown.

\subsubsection{Guided cost learning}
\label{sec:gcl}

Guided cost learning introduces an iterative sample-based method for estimating
$Z$ in the MaxEnt IRL formulation, and can scale to high-dimensional state and action spaces and nonlinear cost functions~\cite{fla-gcl-16}.
The algorithm estimates $Z$ by training a new sampling distribution $\qt$ and using importance sampling:
\begin{align*}
\loss{cost}{\params} &= \E{\traj \sim p}{-\log \pt}
= \E{\traj \sim p}{\c\of{\traj}} + \log Z \\
&= \E{\traj \sim p}{\c\of{\traj}} + \logo{\E{\traj \sim q}{\frac {\expc}{\qt}}}.
\end{align*}
Guided cost learning alternates
between optimizing $c_{\theta}$ using this estimate,
and optimizing $\qt$ to minimize the variance of the importance sampling estimate.

\newcommand{\estimatedp}{\widetilde{p}\of{\traj}}


The optimal importance sampling distribution for estimating the partition function $\int \exp(-\cost_\params(\tau)) d\tau$ is
$\qt \propto |\exp(-\cost_\params(\tau))| = \exp(-\cost_\params(\tau))$. During guided cost learning, the sampling policy $\qt$ is updated to match this distribution by
minimizing the KL divergence between $\qt$ and $\frac 1Z \exp(-\cost_\params(\tau))$, or equivalently minimizing the learned cost and maximizing entropy.
\begin{align}\label{policyloss}
\loss{sampler}{q} = \E{\traj \sim q}{\c\of{\traj}} + \E{\traj \sim q}{\log \qt}
\end{align}
Conveniently, this optimal sampling distribution is the demonstration distribution for the true cost function.
Thus, this training procedure results in both a learned cost function, characterizing the demonstration distribution, and a learned policy $\qt$, capable of generating samples from the demonstration distribution.

This importance sampling estimate
can have very high variance if the sampling distribution $q$
fails to cover some trajectories $\traj$ with high values of $\expc$. Since the demonstrations will have low cost (as a result of the IRL objective),
we can address this coverage problem by mixing the demonstration data samples with the generated samples.
Let $\mix = \frac 12 p + \frac 12 q$
be the mixture distribution over trajectory roll-outs.
Let $\estimatedp$ be a rough estimate for the density of the demonstrations; for example we could use the current model $p_{\params}$, or we could use a simpler density model trained using another method.
Guided cost learning uses $\mix$ for importance sampling\footnote{In RL settings,
where generating samples requires executing a policy in the real world, such as in robotics, old samples from old generators are typically retained for efficiency. In this case, the density $q$ can be computed using a fusion distribution over the past generator densities.}, with $\frac 12 \estimatedp + \frac 12 \qt$ as the importance weights:
\vspace{-0.2cm}
\begin{align*}
\loss{cost}{\params}
&= \E{\traj \sim p}{\c\of{\traj}} + \logo{\E{\traj \sim \mix}{\frac {\expc}{\frac 12 \estimatedp +  \frac 12 \qt}}},
\end{align*}

\subsection{Direct Maximum Likelihood and Behavioral Cloning}

A simple approach to imitation learning and generative modeling is to train
a generator or policy to output a distribution over the data, without learning a discriminator or energy function. For tractability, the data distribution is typically factorized using a directed graphical model or Bayesian network.
In the field of generative modeling, this approach has most commonly been applied to 
speech and language generation tasks~\cite{wscl-gntm-16,odzs-wavenet-16}, but has also been applied to image generation~\cite{vkk-pixelrnn-16}.
Like most EBMs, these models are trained by maximizing the likelihood of the observed data points.

When a generative model does not have the capacity to represent the entire data distribution, maximizing likelihood directly will lead to a moment-matching distribution that tries to ``cover'' all of the modes, leading to a solution that puts much of its mass in parts of the space that have negligible probability under the true distribution. In many scenarios, it is preferable to instead produce only realistic, highly probable samples, by ``filling in'' as many modes as possible, at the trade-off of lower diversity. Since EBMs are also trained with maximum likelihood, the energy function in an EBM will exhibit the same moment-matching behavior when it has limited capacity.
However, designing a flexible energy function to represent a distribution's density function is generally much easier than designing a tractable generator with the same flexibility, that can to generate samples without a complex iterative inference procedure. Moreover, once we have a trained energy function, the generator is trained to be mode-seeking, by minimizing the KL divergence between the generator's distribution and the distribution induced by the energy function. As a result, even if the generator has the same capacity as a generative model trained with direct maximum likelihood, the generator trained with an EBM will exhibit mode-seeking behavior as long as the energy function is more flexible than the generator.
Of course, this phenomenon is often achieved at the cost of tractability, as generating samples from an energy function requires training a generator which, in the case of IRL, is forward policy optimization. 


In sequential decision-making domains, using direct maximum likelihood is known as behavioral cloning, where the policy is trained with supervised learning to match the actions of the demonstrating agent, conditioned on the corresponding observations.  While this approach is simple and often effective for small problems, the moment-matching behavior of direct maximum likelihood can produce particularly ineffective trajectories because of compounding errors. When the policy makes a small mistake, it deviates from the state distribution seen during training, making it more likely to make a mistake again. This issue compounds and eventually, the agent reaches a state far from the training distribution and makes a catastrophic error~\cite{rgb-rilsp-11}. Generative modeling also faces this issue when generating variables sequentially. A popular approach for handling this  involves incrementally sampling more from the model and drawing less from the data distribution during training~\cite{rgb-rilsp-11}. This requires that the true data distribution can be sampled from during training, corresponding to a human or algorithmic expert. Bengio et al. proposed an approximate solution, termed scheduled sampling, that does not require querying the data distribution~\cite{bvjs-schedsampl-15}. However, while these approaches alleviate the issue, they do not solve it completely.

\section{GANs and IRL}

We now show how generative adversarial modeling has implicitly been applied to the setting of inverse reinforcement learning, where the data-to-be-modeled is a set of expert demonstrations. The derivation requires a particular form of discriminator, which we discuss first in Section~\ref{sec:discr}. After making this modification to the discriminator, we obtain an algorithm for IRL, as we show in
Section~\ref{sec:equivalence}, where
 the discriminator involves the learned cost and the generator represents the policy.

\subsection{A special form of discriminator}
\label{sec:discr}

For a fixed generator with a [typically unknown] density $\qt$, the optimal discriminator is the following~\cite{gpmxw-gan-14}:
\vspace{-0.1cm}
\begin{equation}
\label{eq:opt}
D^*(\traj) = \frac{\demodistr}{\demodistr + \qt},
\end{equation}
where $\demodistr$ is the actual distribution of the data.
\newcommand{\discriminator}{D_{\theta}}

In the traditional GAN algorithm, the discriminator is trained to directly output this value.
When the generator density $\qt$ can be evaluated,
the traditional GAN discriminator can be modified
to incorporate this density information. Instead of having the discriminator estimate the value of Equation~\ref{eq:opt} directly, it can be used to estimate $\demodistr$, filling in the value of $\qt$ with its known value. In this case, the new form of the discriminator $\discriminator$ with parameters $\theta$ is
$$
\discriminator(\traj) = \frac{\tilde{p}_\theta(\traj)}{\tilde{p}_\theta(\traj) + \qt}.
$$


In order to make the connection to MaxEnt IRL, we also replace the estimated data density with the Boltzmann distribution. As in MaxEnt IRL, we write the energy function as $\cost_\params$ to designate the learned cost. Now the discriminator's output is:
\[\discriminator\of{\traj} = \frac{\frac 1Z \exp\of{-\c\of{\traj}}}{\mixeddensity{\traj}}.\] 
 
The resulting architecture for the discriminator is very similar to a typical model for binary classification, with a sigmoid as the final layer and $\log Z$ as the bias of the sigmoid.
We have adjusted the architecture only by subtracting $\log \qt$ from the input to the sigmoid.
This modest change allows the optimal discriminator to be completely independent of the generator:
the discriminator is optimal when $\frac 1Z \exp\of{-\c\of{\traj}} = p\of{\traj}$.
Independence between the generator and the optimal discriminator
may significantly improve the stability of training.

This change is very simple to implement and is applicable in any setting where the density $\qt$ can be cheaply evaluated.
Of course this is precisely the case where we could directly maximize likelihood, and we might wonder whether it is worth the additional complexity of GAN training.
But the experience of researchers in IRL has shown that maximizing log likelihood directly is not always the most effective way to learn complex behaviors, even when it is possible to implement.
As we will show, there is a precise equivalence between MaxEnt IRL and this type of GAN, suggesting that the same phenomenon may occur in other domains: GAN training may provide advantages even when it would be possible to maximize likelihood directly.


\subsection{Equivalence between generative adversarial networks and guided cost learning}
\label{sec:equivalence}

In this section, we show that GANs, when applied to IRL problems, optimize the same objective as MaxEnt IRL, and in fact the variant of GANs described in the previous section is precisely equivalent to guided cost learning.

Recall that the discriminator's loss is equal to
\begin{align*}
\loss{discriminator}{\discriminator}
&=  \E{\traj \sim p}{- \log{\discriminator(\traj)}} + \E{\traj \sim q}{- \logo{1-\discriminator(\traj)}}\\
&=  \E{\traj \sim p} {- \log \frac{\frac 1Z \expc}{\mixeddensity{\traj}}} + \E{\traj \sim q}{- \log \frac{q\of{\traj}}{\mixeddensity{\traj}}} \\
 \end{align*}
In maximum entropy IRL, the log-likelihood objective is:
\vspace{-0.1cm}
\begin{align}\label{restatediocloss}
\loss{cost}{\theta} &=  \E{\traj \sim p}{\c\of{\traj}} + \logo{ \E{\traj \sim \frac 12 p + \frac 12 q}{\frac {\expc}{\frac 12 \estimatedp + \frac 12 \qt}}} \\
&= \E{\traj \sim p}{\c\of{\traj}} + \logo{\E{\traj \sim \mix}{\frac {\expc}{\frac 1{2Z} \expc + \frac 12 \qt}}},
\end{align}
where we have substituted $\estimatedp = \pt = \frac 1Z \expc$,
i.e. we are using the current model to estimate the importance weights.

We will establish the following facts,
which together imply that GANs optimize precisely the MaxEnt IRL problem:
\vspace{-0.1cm}
\begin{enumerate}
\item The value of $Z$ which minimizes the discriminator's loss is an importance-sampling estimator for the partition function, as described in Section~\ref{sec:gcl}.
\item For this value of $Z$, the derivative of the discriminator's loss with respect to $\theta$ is equal to the derivative of the MaxEnt IRL objective.
\item The generator's loss is exactly equal to the cost $c_{\theta}$ minus the entropy of $\qt$, i.e. the MaxEnt policy loss defined in Equation~\ref{policyloss} in Section~\ref{sec:gcl}.
\end{enumerate}
\vspace{-0.1cm}

Recall that $\mix$ is the mixture distribution between $p$ and $q$.
Write $\mut = \frac 1{2Z} \expc + \frac 12 \qt$.
Note that when $\theta$ and $Z$ are optimized, 
$\frac 1Z \expc$ is an estimate for the density of $\demodistr$,
and hence $\mut$ is an estimate for the density of $\mix$.

\subsubsection{$Z$ estimates the partition function}
We can compute the discriminator's loss:
\vspace{-0.1cm}
\begin{align}
\loss{discriminator}{\discriminator}
 =& \E{\traj \sim p} {- \log \frac{\frac 1Z \expc}{\mut}}
 + \E{\traj \sim q}{- \log \frac{q\of{\traj}}{\mut}} \\
= & \log{Z} + \E{\traj\sim p}{\c\of{\traj}}
+ \E{\traj \sim p}{\log \mut}
- \E{\traj\sim q}{\log \qt}
+ \E{\traj \sim q}{\log \mut} \\
= & \log{Z}
+ \E{\traj \sim p}{\c\of{\traj}}
- \E{\traj\sim q}{\log \qt}
+ 2\E{\traj \sim \mu}{\log \mut}. \label{discriminatorloss}
\end{align}
Only the first and last terms depend on $Z$.
At the minimizing value of $Z$,
the derivative of these term with respect to $Z$ will be zero:
\vspace{-0.2cm}
\begin{align*}
\partial_Z \loss{discriminator}{\discriminator} &= 0 \\
\frac 1{Z} &= \E{\traj \sim \mix}{\frac{\frac{1}{Z^2} \expc}{\mut}} \\
Z &= \E{\traj \sim \mix}{\frac {\expc}{\mut}}.
\end{align*}
Thus the minimizing $Z$ is precisely the importance sampling estimate of the partition function in Equation~\ref{restatediocloss}.

\subsubsection{$\c$ optimizes the IRL objective}

We return to the discriminator's loss as computed in
Equation~\ref{discriminatorloss}, and consider the derivative with respect to the parameters $\theta$.
We will show that this is exactly the same as the derivative of the IRL objective.

Only the second and fourth terms in the sum depend on $\theta$.
When we differentiate those terms we obtain:
\vspace{-0.1cm}
\newcommand{\dc}{\partial_{\theta}\c\of{\traj}}
\begin{align*}
\partial_{\theta} \loss{discriminator}{\discriminator} =
\E{\traj \sim p}{\dc}
- \E{\traj \sim \mix}{\frac {\frac 1Z \expc \dc}{\mut}}
\end{align*}

On the other hand, when we differentiate the MaxEnt IRL objective, we obtain:
\begin{align*}
    \partial_{\theta} \loss{cost}{\theta}
    &= \E{\traj \sim p}{\dc} + \partial_{\theta} \logo{\E{\traj \sim \mix}{\frac{\expc}{\mut}}} \\
    &= \E{\traj \sim p}{\dc} +  \left( \E{\traj \sim \mix}{ \frac{- \expc \dc}{\mut}} \middle/ \E{\traj \sim \mix}{\frac{\expc}{\mut}} \right) \\
    &= \E{\traj \sim p}{\dc}
- \E{\traj \sim \mix}{\frac {\frac 1Z \expc \dc}{\mut}} \\
    &= \partial_{\theta} \loss{discriminator}{\discriminator}.
\end{align*}
In the third equality, we used the definition of $Z$ as an importance sampling estimate.
Note that in the second equality, we have treated $\mut$
as a constant rather than as a quantity that depends on $\theta$. This is because the IRL optimization is minimizing $\log Z = \log \sum_{\traj} \expc$ and using $\mut$ as the weights for an importance sampling estimator of $Z$.
For this purpose we do not want to differentiate through the importance weights.

\vspace{-0.1cm}
\subsection{The generator optimizes the MaxEnt IRL objective}

Finally, we compute the generator's loss:
\begin{align*}
\loss{generator}{q}
&= \E{\traj \sim q}{\logo{1 - D\of{\traj}} - \logo{D\of{\traj}}} \\
&= \E{\traj \sim q}{\log \frac{\qt}{\mut} - \log \frac{\frac 1Z \expc}{\mut}} \\
&= \E{\traj \sim q}{\log \qt + \log Z  + \c\of{\traj}} \\
&= \log Z + \E{\traj \sim q}{\c\of{\traj}} + \E{\traj \sim q}{\log \qt}
= \log Z + \loss{sampler}{q}.
\end{align*}
The term $\log Z$ is a parameter of the discriminator
that is held fixed while optimizing the generator,
 this loss is exactly equivalent
 the sampler loss from MaxEnt IRL, defined in Equation~\ref{policyloss}.

\subsection{Discussion}

There are many apparent differences between MaxEnt IRL and the GAN optimization problem.
But, we have shown that after making a single key change---using a generator $\qt$ for which densities can be evaluated efficiently, and incorporating this information into the discriminator in a natural way---generative adversarial networks can be viewed as a sample-based algorithm for the MaxEnt IRL problem.
By connecting GANs to the empirical literature on inverse reinforcement learning~\cite{fla-gcl-16}, this demonstrates that GAN training can improve the quality of samples even when the generator's density can be evaluated exactly.
By generalizing this connection, we can derive a new adversarial training strategy for energy-based models, which we discuss in the next section.
\vspace{-0.1cm}

\section{GANs for training EBMs}
\vspace{-0.1cm}

Now that we have highlighted the connection between GANs and guided cost learning, the application of GANs to EBMs follows directly. As discussed in Section~\ref{sec:ebms}, the primary challenge in training  EBMs is estimating the partition function, which is done by approximately sampling from the distribution induced by the energy $E_\params$. Two recent papers have proposed to use adversarial training to derive fast estimates of the partition function~\cite{kb-ddgm-16,zml-ebgan-16}. In particular, these methods alternate between training a generator to produce samples with minimal energy $E_\params(\gansamp)$, and optimizing the parameters of the energy function using the samples to estimate the partition function.

\newcommand{\estimatedpebm}{\widetilde{p}\of{\gansamp}}
\newcommand{\Ex}{E_{\params}\of{\gansamp}}
\newcommand{\expE}{\expo{-\Ex}}
\newcommand{\qx}{q(\gansamp)}

When the density of the generator is available, however, we can derive an unbiased estimate of the partition function as
$$
Z = \E{\gansamp \sim \mix}{\frac {\expE}{\frac 12 \estimatedpebm +  \frac 12 \qx}}
$$
where $\mix$ denotes an equal mixture of generated and real data points, $q(\gansamp)$ denotes the density under the generator, and $\estimatedpebm$ denotes an estimate for the data density. 

This gives a loss function
\vspace{-0.05cm}
\begin{align*}
\loss{energy}{\theta}
&= \E{\gansamp \sim p}{-\log \pebm} \\
&= \E{\gansamp \sim p}{-\Ex} - \logo{\E{\gansamp \sim \mu}{\frac{\expE}{\frac 12 \estimatedpebm + \frac 12 \qx}}}.
\end{align*}

As before, the generator is updated to minimize energy and maximize entropy:
$$
\loss{generator}{q} = \E{\gansamp \sim q}{E_\params \of{\gansamp}} + \E{\gansamp \sim q}{\log \qx}
$$

If we set $\estimatedpebm = \pebm$, the resulting model is a special case of a GAN which is straightforward to implement. The discriminator's output is $\sigma\of{E_{\theta}\of{\gansamp} - \log q\of{\gansamp}}$, where $\sigma$ is a sigmoid with a trainable bias. The discriminator's loss is the log probability and the generator's loss is the discriminator's log odds, as defined in Section~\ref{gans}.

Kim \& Bengio proposed a similar energy-based model
for generative image modeling, but did not assume they could compute the generator's density~\cite{kb-ddgm-16}. As a result, they do not use importance weights, and work with a biased estimator of the partition function which converges to the true partition function when the generator correctly samples from the energy-based model. In contrast, by using the generator density, we can get an unbiased estimate of the partition function that does not rely on any assumptions about the generator. Thus, even if the generator cannot learn to sample exactly from the data distribution, our training procedure is consistent.



Zhao et al. also proposed an energy-based GAN model with an autoencoder discriminator where the energy is given by the mean-squared error between the data example
(generated or real) 
and the discriminator's reconstruction~\cite{zml-ebgan-16}. The energy function is optimized with a margin loss, and the generator is trained to minimize energy. This method also did not use the form of discriminator presented above.
An interesting direction for future exploration is to consider combining the GAN training algorithm discussed here with an objective other than log-likelihood, such as one used with EBMs~\cite{lchr-ebl-06} or different $f$-divergences used with GANs~\cite{nct-fgan-16}.



\vspace{-0.15cm}
\section{Related Work}
\vspace{-0.15cm}

Ho et al.~\cite{hge-mfil-16,he-gail-16} previously presented a GAN-like algorithm for imitation learning, where the goal is to recover a policy that matches the expert demonstrations.
The proposed algorithm, called generative adversarial imitation learning (GAIL), has an adversarial structure. 
The analysis in this paper provides additional insight into what GAIL is doing. As discussed above, GANs are optimizing the same objective as MaxEnt IRL. Thus, the GAIL policy is being trained to optimize a cost learned through MaxEnt IRL. Unlike guided cost learning~\cite{fla-gcl-16}, however, Ho \& Ermon use the typical unconstrained form of the discriminator~\cite{he-gail-16} and do not use the generator's density.
In this case, the cost function remains implicit within the discriminator and cannot be recovered. Hence, in GAIL, the discriminator is discarded and the policy is the end result.

Bachman \& Precup~\cite{bp-dgsdm-15} suggested that data generation can be converted into a sequential decision-making problem and solved with a reinforcement learning method. Several recent works have proposed methods for merging maximum likelihood objectives and known reward functions for training sequential language generation models and rely on surrogate reward function such as BLEU score or edit distance~\cite{rcaz-slt-16,nbcj-raml-16,bbxg-acsp-16}. In this work, we assume that the reward function is unknown.

Yu et al. proposed to learn a cost function for sequential data generation using GANs, where the cost is defined as the probability of the discriminator classifying the generated sequence as coming from the data distribution~\cite{yzwy-seqgan-16}. The discriminator does not take advantage of the policy's density values, despite the fact that they are known (and are used during pre-training). Their experiments also find that max-likelihood pre-training is crucial for good performance, suggesting that recurrent generators that can't afford such pre-training (e.g. because they don't have densities) are less practical to train.

Pfau \& Vinyals drew a connection between the optimization problems in GANs and actor-critic methods in reinforcement learning, suggesting how ideas for stabilizing training in one domain could be beneficial for the other~\cite{pv-ganac-16}. As the authors point out, these optimization tricks could also be useful for imitation learning algorithms with the same two-level optimization structure.






\vspace{-0.1cm}
\section{Discussion}
\label{conclusion}
\vspace{-0.15cm}

In this work, we showed an equivalence between generative adversarial modeling and an algorithm for performing maximum entropy inverse reinforcement learning.
Our derivation used a special form of discriminator that leverages likelihood values from the generator, leading to an unbiased estimate of the underlying energy function. A natural direction for future work is to experiment with combining deep generators that can provide densities, such as autoregressive models~\cite{lm-nade-11,vkk-pixelrnn-16} or models that use
invertible transformations~\cite{dsb-realnvp-16}, with generative adversarial modeling. Such an approach may provide more stable training, better generators, and wider applicability to discrete problems such as language.

This work also suggests a new algorithm for training energy-based models using generative adversarial networks, that trains a neural network model to sample from the distribution induced by the current energy. This method could reduce the computational challenges of existing MCMC-based solutions.


\section*{Acknowledgments}
\vspace{-0.1cm}
The authors would like to thank Ian Goodfellow and Joan Bruna for insightful discussions.

\bibliographystyle{abbrv}
\bibliography{references}
\end{document}

%% file: defs.tex
\long\def\ignorethis#1{}

\renewcommand{\eqref}[1]{Equation~(\ref{#1})}

\newcommand{\gansamp}{\mathbf{x}}

\newcommand{\params}{\theta}

\newcommand{\cost}{c}
\newcommand{\state}{\mathbf{x}}
\newcommand{\action}{\mathbf{u}}

\newcommand{\traj}{\tau}

\newcommand{\obj}{\mathcal{L}}

\newcommand{\datadistr}{p(\mathbf{x})}
\newcommand{\demodistr}{p(\tau)}